\documentclass[journal]{IEEEtran}
\usepackage{amsmath,amsfonts}
\usepackage{algorithmic}
\usepackage{algorithm}
\usepackage{array}
\usepackage[caption=false,font=normalsize,labelfont=rm,textfont=rm]{subfig}
\usepackage{textcomp}
\usepackage{tablefootnote}

\usepackage{stfloats}
\usepackage{url}
\usepackage{verbatim}
\usepackage{graphicx}
\usepackage{cite}
\usepackage{bbding}
\usepackage{colortbl}
\usepackage{multirow}
\usepackage{longtable}
\begin{document}

\title{An Aerial Manipulator for Robot-to-robot Torch Relay Task: System Design and Control Scheme}

\author{Guangyu Zhang, Yuqing He, Liying Yang, Chaoxiong Huang, Yanchun Chang, Siliang Li
  \thanks{Manuscript received Sep. 29 2022; accepted Mar. 15, 2023.}
  \thanks{
    G. Zhang, Y. He, L. Yang, C. Huang, Y. Chang, and S. Li are with the State Key Laboratory of Robotics, Shenyang Institute of Automation, Chinese Academy of
    Sciences, Shenyang 110016, China, and also with the Institute for Robotics and Intelligent Manufacturing, Chinese Academy of Sciences, Shenyang 110016, China(zhangguangyu, heyuqing, yangliying@sia.cn).
  }
	
}

\markboth{IEEE Robotics \& Automation Magazine (Accepted Version)}%
{Shell \MakeLowercase{\textit{et al.}}: A Sample Article Using IEEEtran.cls for IEEE Journals}


\maketitle

\begin{abstract}
  Torch relay is an important tradition of the Olympics and heralds the start of the Games. Robots applied in the torch relay activity can not only demonstrate the technological capability of humans to the world but also provide a sight of human lives with robots in the future. This article presents an aerial manipulator designed for the robot-to-robot torch relay task of the Beijing 2022 Winter Olympics. This aerial manipulator system is composed of a quadrotor, a 3 DoF (Degree of Freedom) manipulator, and a monocular camera. This article primarily describes the system design and system control scheme of the aerial manipulator. The experimental results demonstrate that it can complete robot-to-robot torch relay task under the guidance of vision in the ice and snow field.
\end{abstract}
\begin{IEEEkeywords}
  aerial manipulation, aerial manipulator, aerial robot, robot control
\end{IEEEkeywords}
\section{Introduction}
\begin{figure}[!t]
  \centering
  \includegraphics[width=1\linewidth]{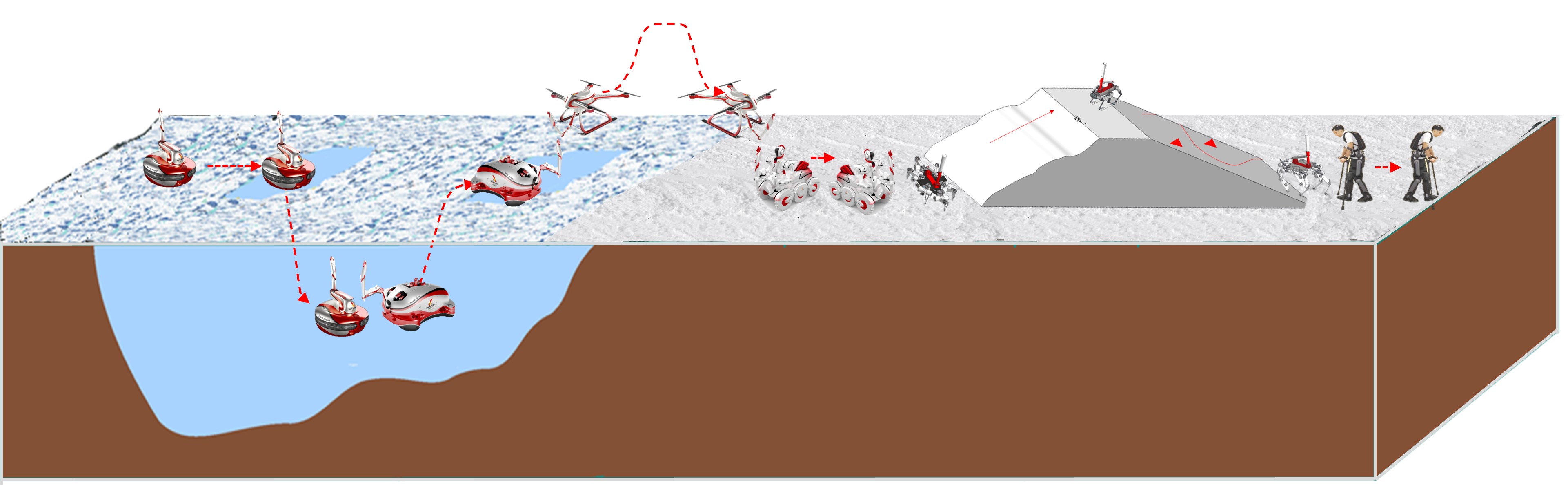}
  \caption{Scenario of Olympic torch relay between robots.}
  \label{fig_1}
\end{figure}

\IEEEPARstart{I}{n} the Beijing 2022 Winter Olympic Games, several types of robots have been designed for the Olympic torch relay activity, including aerial, underwater, hexapod, and other robots\cite{gao_robots_2022}. The designed scenario is  that the Olympic torch would be relayed between robots in the ice and snow field, as shown in Fig. \ref{fig_1}. It is a chance for the robotics community to show robotic technologies to the world. As shown in the designed scenario, the key for a robot to complete the robot-to-robot torch relay task is that it should be able to light its torch actively from another one in the harsh environment. The designed aerial robot for the torch relay task is a robot known as aerial manipulator, which is mainly composed of a rotor wing UAV (Unmanned Aerial Vehicle) and a robotic arm. With this system structure, the aerial robot can fly and manipulate and be qualified for the torch relay task.
\subsection{Related works}

The aerial manipulator is a new type of aerial robot that has come into being in the last decade\cite{ollero_past_2022}. The capability of flying and manipulation makes it have wide application prospects, such as industrial inspection and maintenance\cite{ollero_aeroarms_2018}. Existing literature about aerial manipulator mainly focuses on  the system design, motion control, and manipulation task control\cite{ollero_past_2022}

In the studies of aerial manipulator system design, different types of manipulators and UAVs are used to satisfy the requirements of different applications. 
For the pick-and-place task, a strong payload capacity and a large workspace are needed. Thus, aerial manipulators composed of a helicopter/multirotor and a serial link robotic arm are designed for stationary or moving object grasping \cite{kondak2014aerial, zhang_grasp_2018}.
For the contact inspection task, the end-effector needs to interact with the environment physically. Therefore, to improve the safety of contact manipulation, the compliant manipulator based aerial manipulator system is designed \cite{bartelds_compliant_2016, suarez_lightweight_2018}.
Moreover, some tasks, such as the push-and-slide task, need the aerial manipulator to provide dexterous active interaction force. Thus, the multi-directional and omni-directional thrust multirotor are used in the system design to satisfy this capacity requirement \cite{tognon_truly-redundant_2019, bodie_dynamic_2021}.
In addition, to improve the aerial manipulator system flight stability, the parallel manipulator also appeared in the system design due to the small inertia of its moving parts \cite{chermprayong_integrated_2019, bodie_dynamic_2021}.

In the studies of aerial manipulator motion control, centralized and decentralized control strategy are used in the controller design.
For the centralized control strategy, the aerial manipulator is considered as a unique entity, and one controller is designed to simultaneously stabilize all the states of the UAV and manipulator. Based on the full dynamic model, the model based control approaches are used in this control scheme, such as dynamics compensation based control \cite{bodie_dynamic_2021}, back-stepping\cite{yang2014dynamics} and MPC (Model Predictive Control) \cite{Lee2020aerial}.
For the decentralized control strategy, two separate controllers are designed to control the UAV flight and the end-effector operation, respectively. At the same time, the dynamic coupling between them is considered as an external disturbance to each other. In this control scheme, DoB (Disturbance Observer Based) control\cite{chen2022adaptive}, robust control \cite{zhang_robust_2020} and inverse kinematics based control \cite{zhang_grasp_2018, chermprayong_integrated_2019} methods are used.

To perform a complicated manipulation task, such as the push-and-slide and peg-in-hole task, the aerial manipulator needs to contact the target object and act force and moment on it. Thus, in the studies of aerial manipulator manipulation task control, 
hybrid position and force control \cite{tzoumanikas_RSS_2020}, impedance control \cite{suarez_physical-virtual_2018}, and vision-based impedance control \cite{lippiello_image-based_2018, hu_vision-based_2022} methods are used to satisfy requirements of the complicated aerial manipulation task.

\subsection{Contributions}
Undoubtedly, an aerial manipulator is very suited to perform the robot-to-robot torch relay task. 
However, there are still some challenges to be overcome.
Firstly, the complicated dynamic coupling between the UAV and the manipulator makes it difficult to control an aerial manipulator with high precision.
Moreover, in the ice and snow field, the sunlight change, ice and snow flash, and wind disturbance will seriously damage the target object localization and the system control performance. 
Finally, when the aerial robot is performing the torch lighting operation with the underwater robot, the target torch will be floating due to the wave disturbance, which makes it harder to light the torch. Although in our previous work\cite{zhang_grasp_2018}, the aerial manipulator has completed a target grasping task,
it is realized indoors, which depends on the motion capture system and without other disturbances from the environment.
This paper primarily presents the system design and control method of the aerial robot for the torch relay task.
The main contributions of this article are as follows:
\begin{itemize}
  \item{An aerial manipulator system was developed for the Beijing 2022 Winter Olympics torch relay. It can complete the torch relay task with other robots autonomously in the ice and snow field.}
  \item{Successful manipulating a floating target object by an aerial manipulator outdoors. To the best of our knowledge, this is the first time for an aerial manipulator system to complete such a task.}
\end{itemize}
\section{System design}
The overall aim of the system design is to complete the torch relay task autonomously in the ice and snow field by an aerial manipulator. Therefore, the task requirements, system reliability, and environmental adaptability are mainly considered in the design of our platform. Furthermore, to make the robot more impressive to the public, the artistic quality and "Olympic elements" should be taken into account in the robot's appearance design, which is suggested by the Beijing Winter Olympic Organizing Committee. In this section, the appearance design, hardware components, and software architecture of our aerial manipulator are presented.
\subsection{Hardware design}
\subsubsection{Morphology}
For the state of the art in aerial manipulator system, there are four different morphologies according to the composing UAV and manipulator, as shown in \mbox{Table \ref{tab_1}}. Every morphology has its characteristics and suits different tasks. It is no doubt that the torch flame burning will be threatened by the aerial robot's rotor airflow. In order to make the aerial robot more suited for the torch relay task, the main guiding principle in the system design is to make the torch away from the rotor airflow as far as possible. Thus, the rotor airflow and operation workspace range are the main properties that are taken into account when choosing the morphology of the aerial manipulator system. The ideal morphology should be with a large workspace range, weak rotor airflow intensity, and narrow rotor airflow range. As shown in \mbox{Table \ref{tab_1}}, the morphology composed of a multirotor and a serial manipulator is better than others in these properties.
\begin{table}[h]
  \begin{center}
    \caption{Satisfaction of different morphologies}
    \label{tab_1}
    \resizebox{\columnwidth}{!}{
      \begin{tabular}{| c | c | c | c |}
        \hline
        Morphology                                     & Workspace    & Rotor airflow      & Rotor airflow    \\
                                                       & range        &  intensity &  range   \\
        \hline
        {\includegraphics[height=0.95cm]{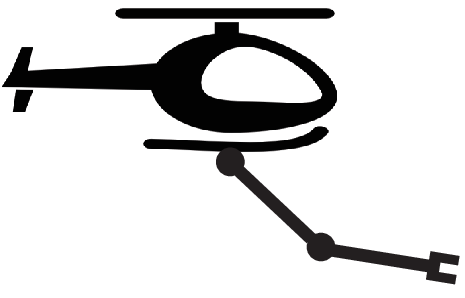}} & \Checkmark   & \XSolidBrush   & \XSolidBrush \\
        Heli-seri\tablefootnote{Helicopter + Serial manipulator}
                                                            &              &                & \\
        \hline
        {\includegraphics[height=1cm]{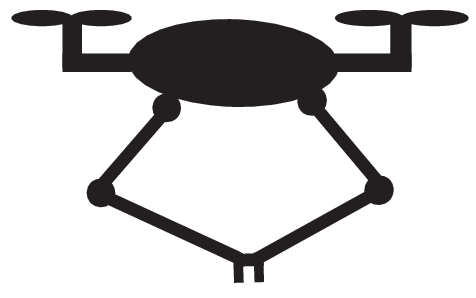}} & \XSolidBrush & \Checkmark     & \Checkmark   \\
         UTmulti-delta\tablefootnote{Uni-directional thrust multirotor + Delta manipulator}
                                                            &              &                & \\
        \hline
        {\includegraphics[height=1cm]{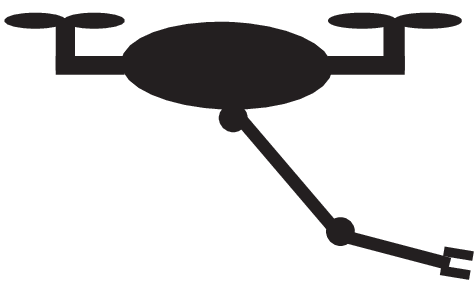}}  & \Checkmark   & \Checkmark     & \Checkmark   \\
         UTmulti-seri\tablefootnote{Uni-directional thrust multirotor + Serial manipulator}
                                                            &              &                & \\
        \hline
                {\includegraphics[height=1cm]{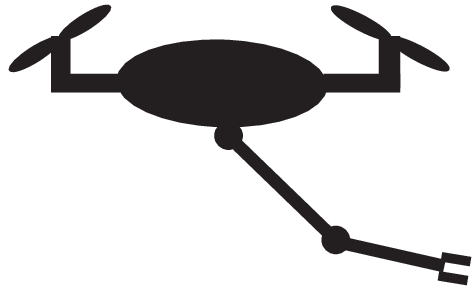}}    & \Checkmark   & \Checkmark     & \XSolidBrush \\
         MTmulti-seri\tablefootnote{Multi-directional thrust multirotor + Serial manipulator\cite{tognon_truly-redundant_2019, bodie_dynamic_2021}} 
                                                            &              &                & \\
        \hline
      \end{tabular}}
  \end{center}
\end{table}
\subsubsection{Appearance}
\begin{figure}[ht]
  \centering
  \subfloat[]{\includegraphics[width=1\linewidth]{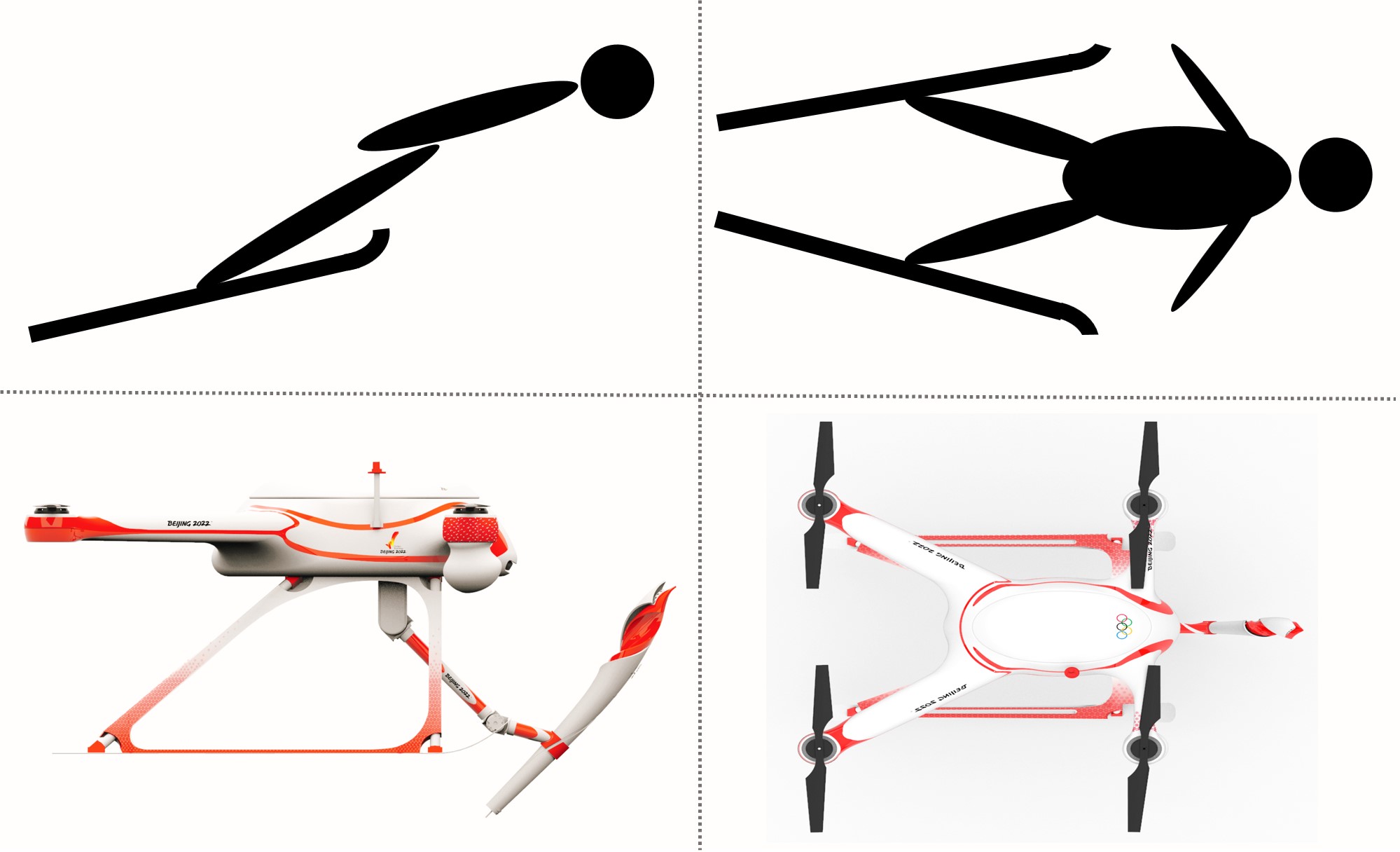}%
    }
  \hfil
  \subfloat[]{\includegraphics[width=1\linewidth]{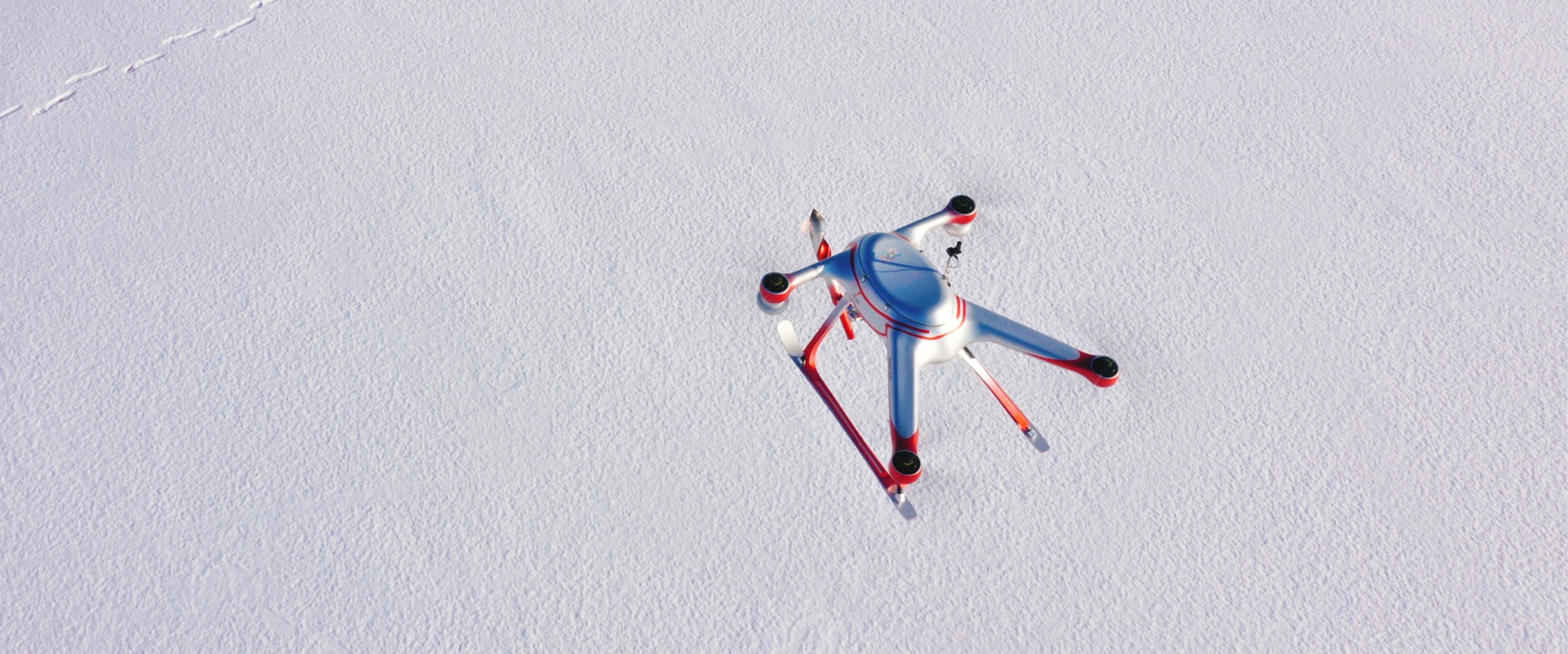}%
    }
  \hfil
  \caption{Illustration of competitor's aerial style of ski jumping and aerial robot's body shape: (a) lateral view and top view; (b) tilt top view.}
  \label{fig_2}
\end{figure}

In the design of the robot's appearance, the Winter Olympic sports elements are integrated. As in Fig. \ref{fig_2}, the aerial robot's main body shape is inspired by the competitor's aerial style of ski jumping, in which competitors can jump and "fly" after skiing down from a ramp, and the aerial robot's landing gear is designed as a ski board. Thus, the aerial robot will look like an athlete competing in ski jumping when flying overhead. To fit the appearance of "Feiyang" torch, which is the torch of the Beijing 2022 Winter Olympics, special decorative patterns are also designed, which makes the torch look like one part of the aerial robot.
\subsubsection{Components}
\begin{figure}[ht]
  \centering
  \subfloat[]{\includegraphics[width=1\linewidth]{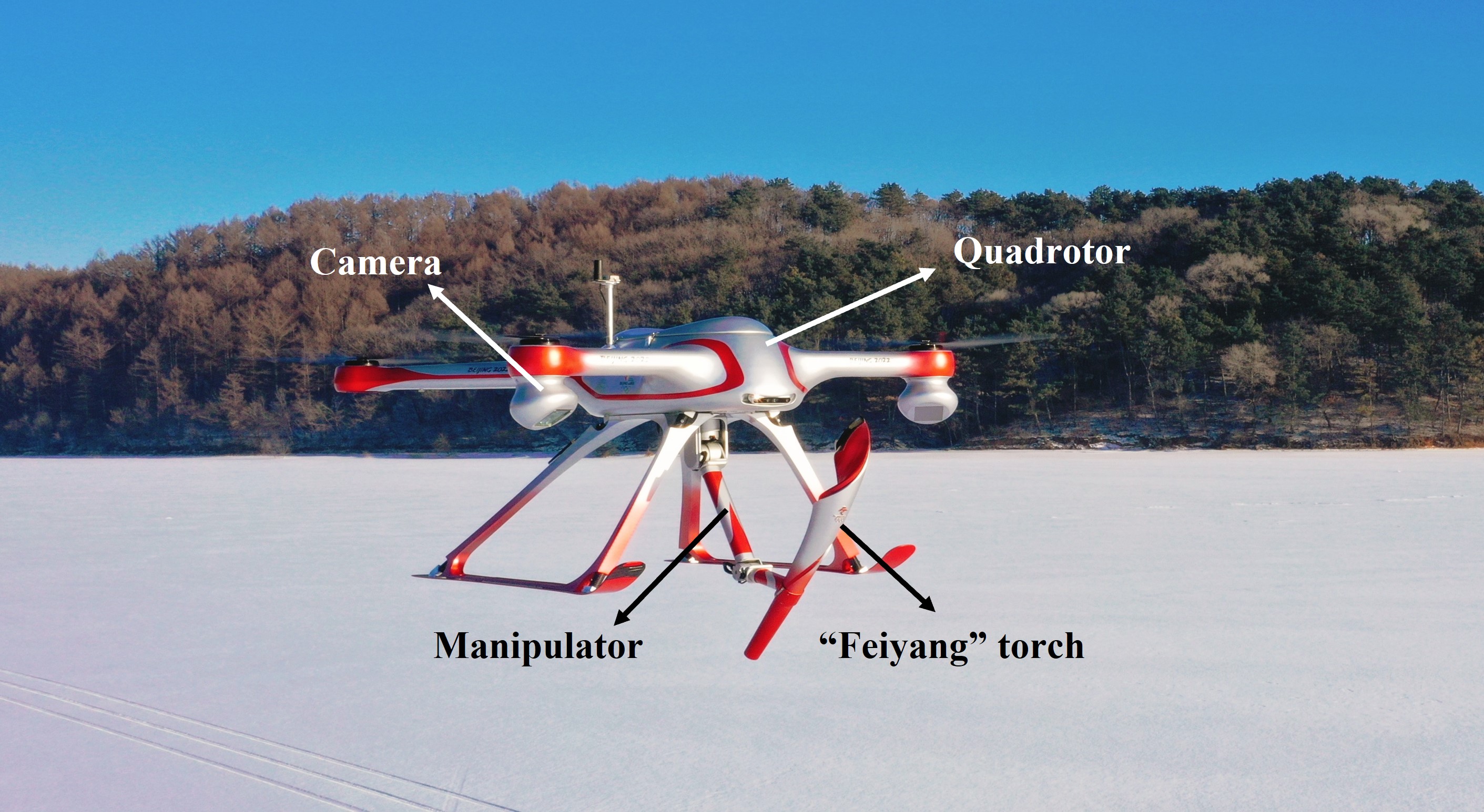}%
  \label{fig_3a}}
  \hfil
  \subfloat[]{\includegraphics[width=0.488\linewidth]{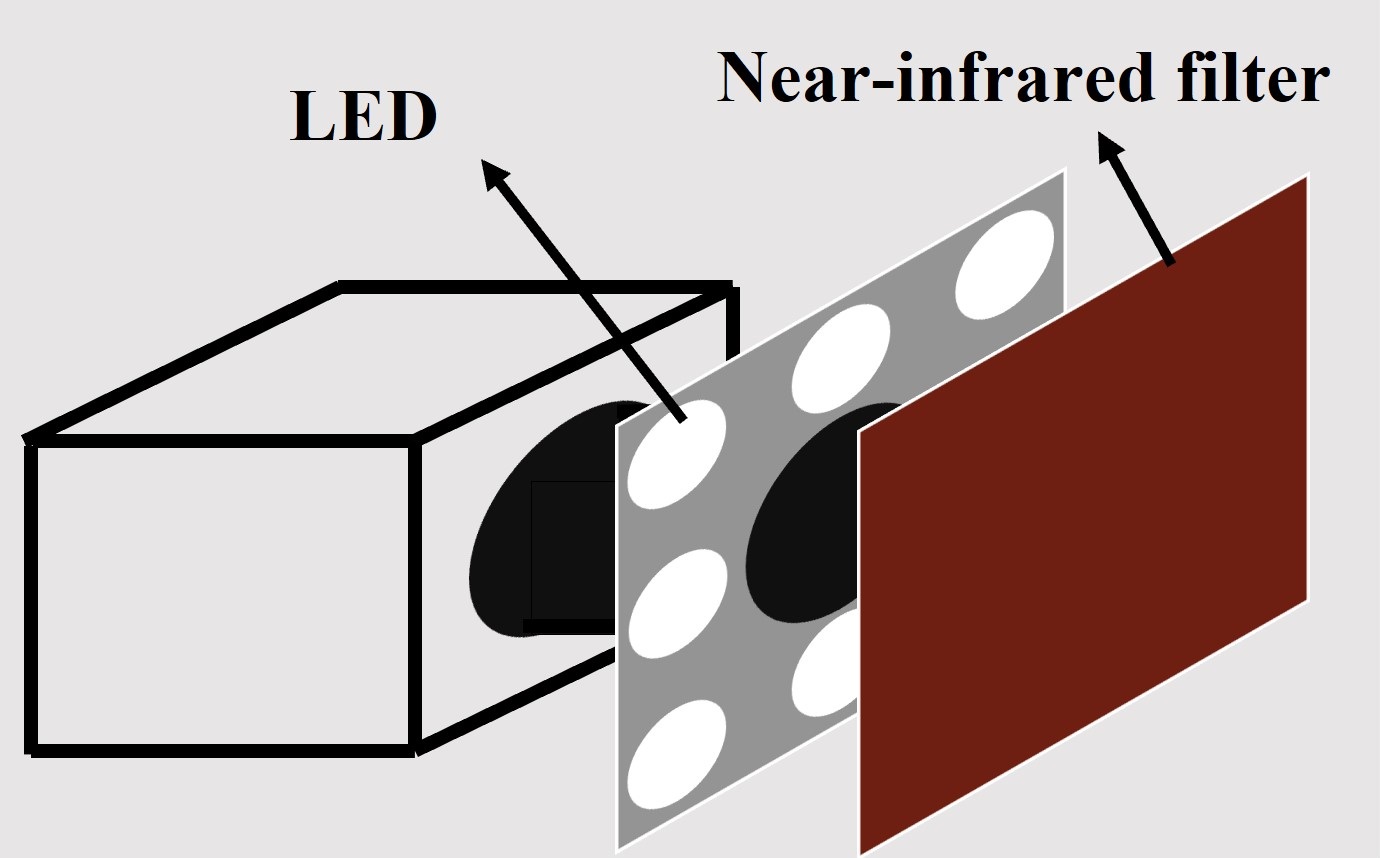}%
  \label{fig_3b}}
  \hfil
  \subfloat[]{\includegraphics[width=0.512\linewidth]{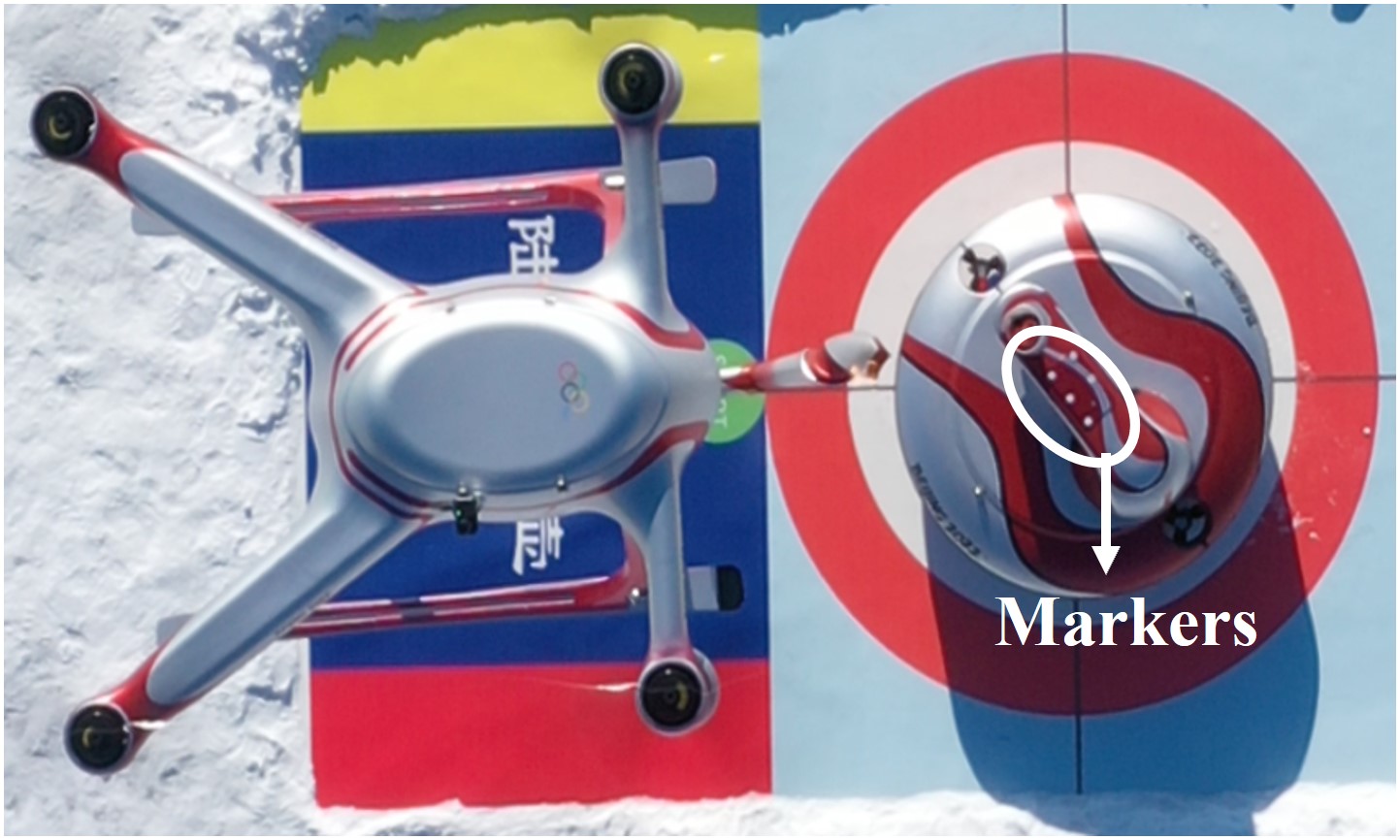}%
  \label{fig_3c}}
  \caption{Aerial manipulator system: (a) system components; (b) structure of vision module; (c) target robot with makers.}
  \label{fig_3}
\end{figure}

To perform the torch relay task autonomously, the aerial robot needs to have the capability of flying, manipulation, pose estimation of target torch, and flame burning state perception, etc. In the overall view, the components of our aerial manipulator system are shown in \mbox{Fig. \ref{fig_3a}}. It mainly consists of a quadrotor, a monocular camera, a manipulator, and a "Feiyang" torch. In the torch, a thermocouple is integrated to sense the flame burning state by the temperature of the flame center. The reason why we chose the quadrotor is that it is more adaptable to the appearance designed in the previous subsection. The manipulator has three joints, so the manipulator can keep the torch end tip in the fire point while the quadrotor is floating under environmental disturbance. The main specifications of the aerial manipulator are listed in \mbox{Table \ref{tab_2}}. 

\begin{table}[H]
  \begin{center}
    \caption{Specification of the aerial robot}
    \label{tab_2}
    \begin{tabular}{| c | c |}
      \hline
      Name                & Value  \\
      \hline
      Length              & 135cm \\
      \hline
      Width               & 105cm \\
      \hline
      Frame height        & 50cm  \\
      \hline
      Total mass          & 35kg  \\
      \hline
      Fight time          & 20min \\
      \hline
      Max. takeoff weight & 40kg  \\
      \hline
      Manipulator length  & 70cm  \\
      \hline
      Manipulator payload & 2kg   \\
      \hline
      Torch length        & 80cm  \\
      \hline
      Torch weight        & 1.5kg \\
      \hline
    \end{tabular}
  \end{center}
\end{table}
To adapt to the ice and snow environment, the aerial robot should have the ability of low temperature resistant. The temperature in the ice and snow field in Beijing can be as low as -20$^{\circ}$C which can make the battery, joint actuators, and other electronic devices unable to work well. To improve low-temperature resistance capacity, an auxiliary heating system is integrated into the aerial robot cabin, where the battery, onboard computer, fight controller, radio transmitter, and all remaining electronic devices are installed. The auxiliary heating system has two heater fans to ensure air circulation in the cabin and keep the temperature above 5$^{\circ}$C. On the other hand, for the devices that can't be installed in the cabin, such as actuator and camera, we select the one with low operating temperature ability, as shown in the \mbox{Table \ref{tab_3}}. All the designs can make the aerial manipulator system work well in harsh outdoor environments.
\begin{table}[H]
  \begin{center}
    \caption{Operation temperature of main devices}
    \label{tab_3}
    \begin{tabular}{| c | c | c |}
      \hline
      Name                           & Model	         & Operating\\
                                     &                 & temperature\\
      \hline
      Flight controller	             & CUAV V5+	       & -20-80$^{\circ}$C\\
      \hline
      Actuator                       & MINITASCA QQD   & \\
      of shoulder roll/pitch joint   &  PR60-36	       & -20-50$^{\circ}$C\\
      \hline
      Actuator                       & MINITASCA QQD   & \\
      of elbow pitch joint	         & NE30-6	         & -20-50$^{\circ}$C\\
      \hline
      Camera                         &IMPER B1310	     & -40-85$^{\circ}$C\\
      \hline
    \end{tabular}
  \end{center}
\end{table}
The primary function of the vision module is to get the pose of the target torch, so that the aerial manipulator can perform torch lighting operation actively with other robots. Object pose estimation based on vision for robot manipulation is a classic research point for industrial robots, and some methods have been proposed, such as RGB-D based and ArUco marker based methods. For our task, the sunlight changes and ice and snow flash make solutions used in the indoor scene unreliable. To make the vision system more reliable in the ice and snow field, as shown in Fig. \ref{fig_3b}, 8 LEDs are integrated around the camera lens, and a near-infrared filter covers them. When the camera takes a picture, the LEDs emit light that passes the near-infrared filter. Thus, suppose the target object is marked by reflective markers. In that case, by reflecting the near-infrared light, the camera can get only the marker feature without any other background, as shown in Fig. \ref{fig_3c}, which means that the environmental light and background will not affect the vision system. 
\subsubsection{Configurations}
\begin{figure}[!t]
  \centering
  \includegraphics[width=1\linewidth]{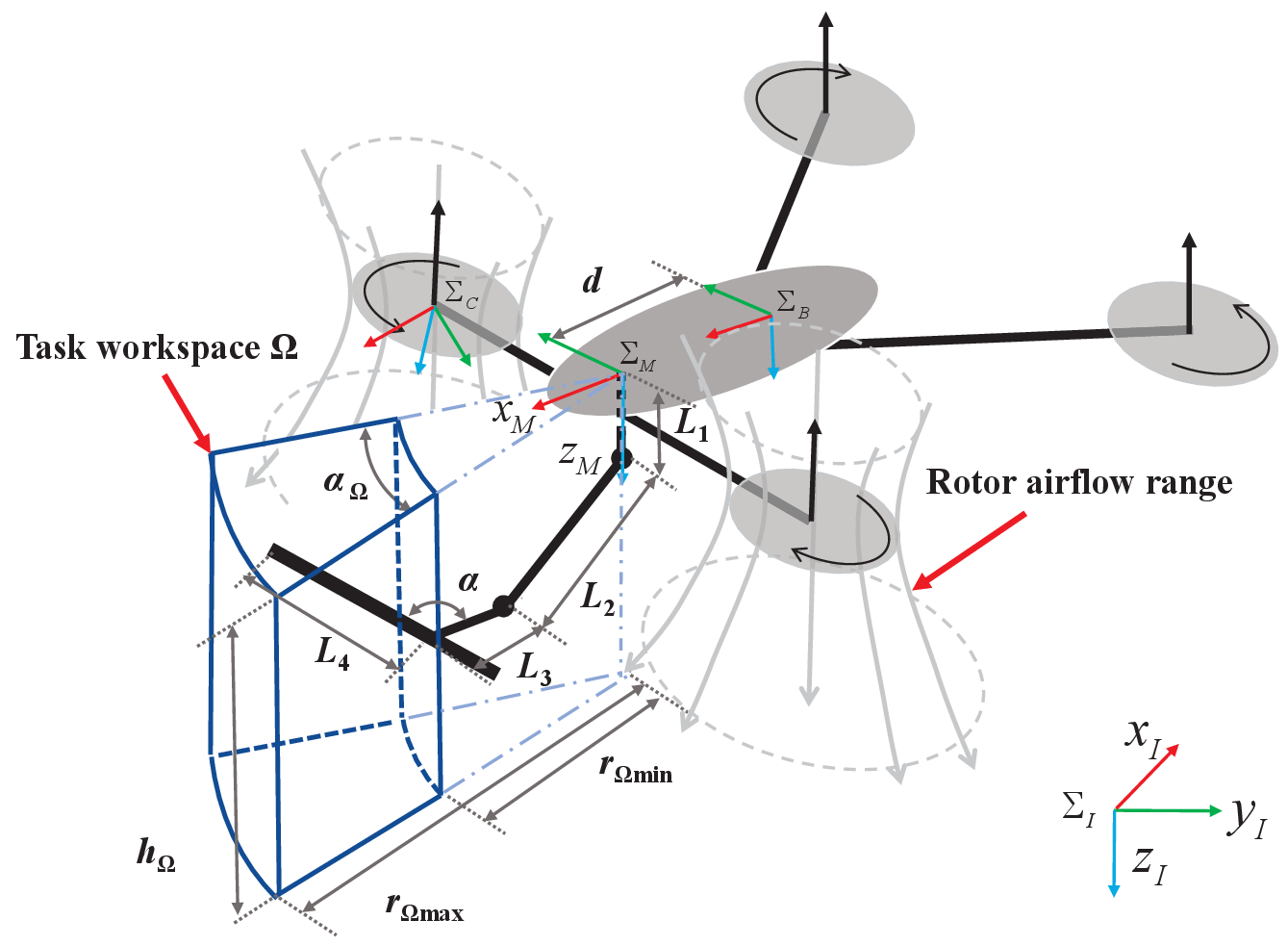}
  \caption{Illustration of the frames, configuration, and task workspace of the aerial manipulator system.}
  \label{fig_4}
\end{figure}

To make the torch flame away from the threat of the rotor airflow, the manipulator is installed in the front of the quadrotor, as shown in Fig. \ref{fig_4}. The camera is installed under the right front rotor and with an orientation toward to the task workspace, so it can keep the target within the field of the camera view when the aerial manipulator is lighting the torch actively. The three rotational joints of the manipulator are shoulder roll, shoulder pitch, and elbow pitch joint, respectively. The torch is fixed on the manipulator, so it can be seen as the end link of the manipulator. There is a fixed intersection angle, denoted by $\alpha$, between the torch and another link, which makes the aerial robot look like a man flying with a torch in hand. The configurations of the aerial manipulator are listed in \mbox{Table \ref{tab_4}}. 
\begin{table}[H]
  \begin{center}
    \caption{Configuration of the manipulator}
    \label{tab_4}
    \begin{tabular}{| c | c | c |}
      \hline
      Symbol     & Description                                  & Value \\
      \hline
      $d$        & Manipulator base offset (mm)                 & 300   \\
      \hline
      $L_1$      & Length of link1 (mm)                         & 100   \\
      \hline
      $L_2$      & Length of link2 (mm)                         & 400   \\
      \hline
      $L_3$      & Length of link3 (mm)                         & 200   \\
      \hline
      $L_4$      & Length of link4 (mm)                         & 530   \\
      \hline
      $\phi_c$   & Pitch angle of camera (°)                    & -30   \\
      \hline
      $\theta_c$ & Yaw angle of camera (°)                      & -60   \\
      \hline
      $\alpha$   & Torch and manipulator intersection angle (°) & 135   \\
      \hline
    \end{tabular}
  \end{center}
\end{table}
\subsubsection{Task workspace}
 In this paper, we use $\Sigma_{\#}$ to denote a right-hand coordinate frame with origin at point $o_{\#}$ and its axes are $x_{\#}$, $y_{\#}$ and $z_{\#}$ which are represented by red, green, and blue vectors, respectively, as shown in Fig. \ref{fig_4}. The inertial 
 frame, quadrotor body-fixed frame, manipulator base frame, and camera frame are denoted by $\Sigma_I$, $\Sigma_B$, $\Sigma_M$, and $\Sigma_C$, respectively. In order to make torch flame have better viewing, a specific task workspace also needs to be defined in front of the aerial manipulator system. As shown in Fig. \ref{fig_4}, the workspace $\Omega$ is a cube space with a fan-shaped cross-section, whose central axis is $z_M$. ${}^Mp_{end}$ denotes the torch endpoint position with respect to $\Sigma_M$, so it will satisfy the following inequalities:
\begin{equation}
  \begin{cases}
    -h_{\Omega } \leq dot(^Mp_{end},z_M)\leq 0                                                                            \\
    r_{\Omega \min} \leq \left\| ^Mp_{end} \right\|\leq r_{\Omega \max }                                                  \\
    -\frac{1}{2}\alpha_\Omega \leq \left\langle x_M, Proj_{xoy}(^Mp_{end}) \right\rangle \le \frac{1}{2}\alpha_{\Omega } \\
  \end{cases}
  \label{eq1}
\end{equation}
where, $h_{\Omega }$ is the height of the task workspace. $\alpha_\Omega$ is the yaw angle of the task workspace, which is the angle of fan-shaped cross-section and is symmetric about the $x_M$. $dot(.)$ is a function of the dot product of two vectors. $\left\langle.\right\rangle$ is a function that gets the angle between two vectors. $Proj_{xoy}(.)$ is a function that projects a vector into $x\text{-} o \text{-}y$ coordinate plane. $r_{\Omega\min}$ and $r_{\Omega\max}$, are the minimum and maximum radius of the fan-shaped section, respectively. 
The specifications of the task workspace are listed in \mbox{Table \ref{tab_5}}.
\begin{table}
  \begin{center}
    \caption{Specification of the task workspace}
    \label{tab_5}
    \begin{tabular}{| c | c | c |}
      \hline
      Symbol            & Description                           & Value \\
      \hline
      $h_\Omega$        & Height of task workspace (mm)         & 700   \\
      \hline
      $\alpha_\Omega$   & Yaw angle of task workspace (°)       & 40    \\
      \hline
      $r_{\Omega \min}$ & Minimum radius of task workspace (mm) & 650   \\
      \hline
      $r_{\Omega \max}$ & Maximum radius of task workspace (mm) & 950   \\
      \hline
    \end{tabular}
  \end{center}
\end{table}
\subsection{Software architecture}
\begin{figure}[!t]
  \centering
  \includegraphics[width=1\linewidth]{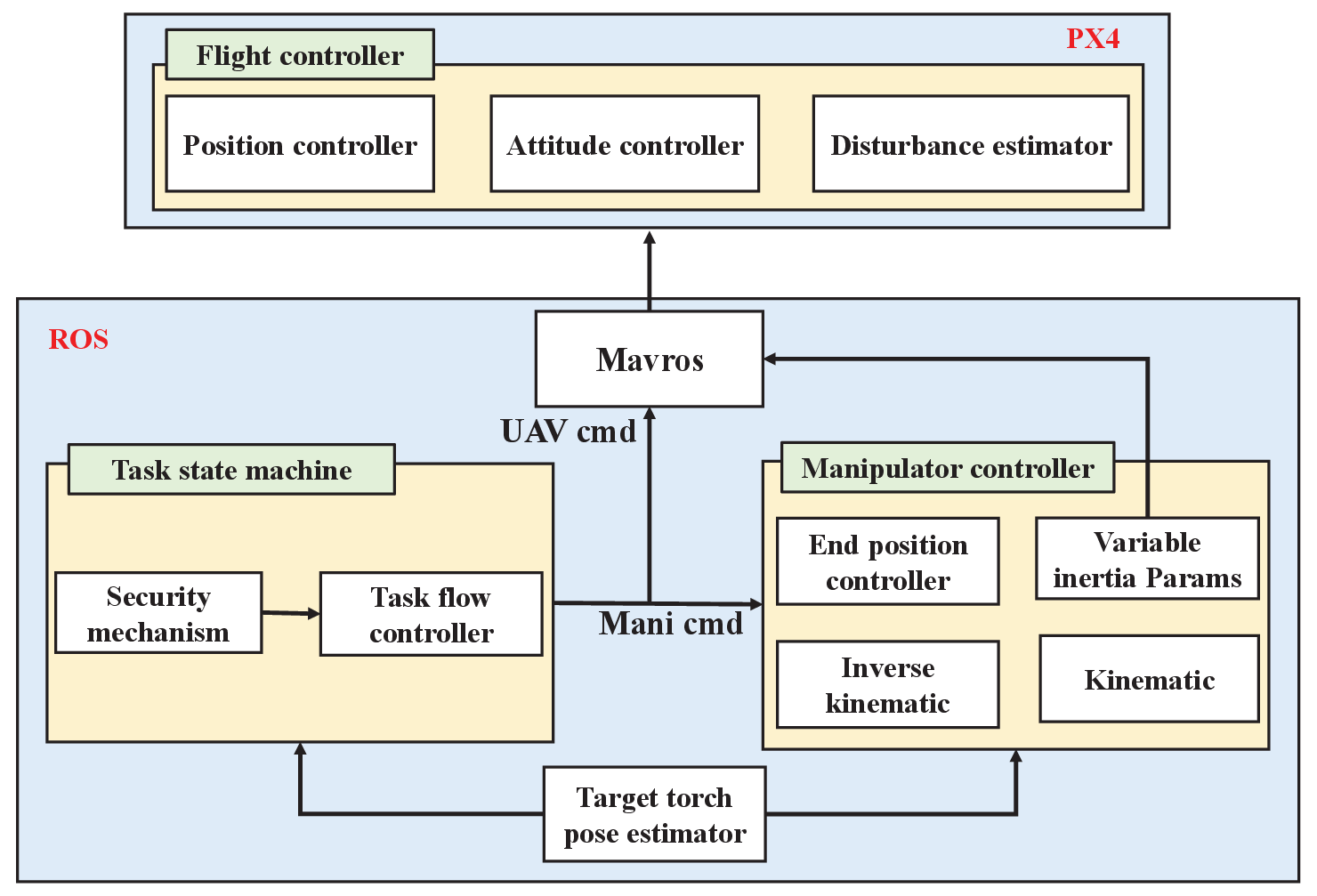}
  \caption{Software architecture.}
  \label{fig_5}
\end{figure}

The software architecture is shown in Fig. \ref{fig_5}, the software implementation is in robotics operating system (ROS) and PX4 firmware\footnote{\url{https://github.com/PX4/PX4-Autopilot/tree/v1.12.0}}. The communication between ROS and PX4 through a ROS node named “mavros”, which can convert MAVLink messages into ROS topics. 
In the PX4 firmware, the flight controller is redeveloped based on our methods, which contains the position controller, attitude controller, and disturbance estimator to estimate the coupling disturbance of the manipulator. Methods of flight controller are detailed in the next Sec.III-B. The modules implemented in the ROS contains the target torch pose estimator, manipulator controller, and task state machine, which send command to the flight controller and manipulator controller based on the task flow. Methods of the manipulator controller are detailed in Sec.III-C.
\section{System control}
In order to perform torch lighting operation, the aerial manipulator needs to hover at one point to keep the target torch in the task workspace and move its torch end tip to the flame of the target torch simultaneously.
Therefore, the performance of steady flight control of the quadrotor and endpoint position control of the manipulator is very important. In this section, system control architecture and methods suitable for the torch relay task are introduced.
\subsection{Control architecture}
\begin{figure*}[!t]
  \centering
  \includegraphics[width=0.9\textwidth]{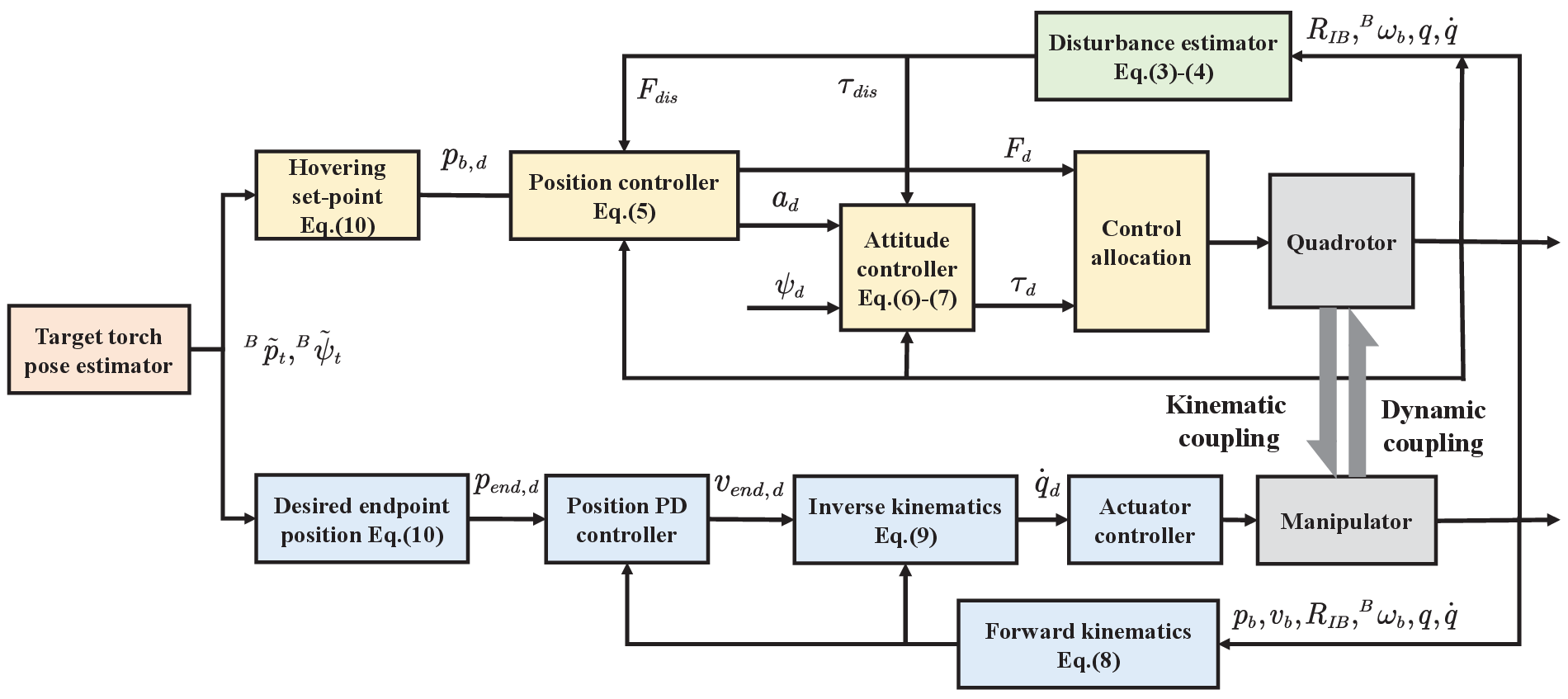}
  \caption{System control architecture.}
  \label{fig_6}
\end{figure*}

The system control architecture is shown in Fig. \ref{fig_6}. The quadrotor and manipulator are controlled separately. In the hovering, the dynamic coupling of the quadrotor effect on the manipulator is very small. Therefore, the dynamic coupling is considered only in the quadrotor controller. For the quadrotor, the dynamic coupling of the manipulator is modeled as disturbance and compensated in the flight controller. For the manipulator, to track the target torch while the quadrotor is floating, the inverse kinematics of the aerial manipulator endpoint is used in the manipulator controller.
\subsection{Quadrotor control}
The quadrotor controller is based on a cascade control structure, as shown in Fig. \ref{fig_6}. In this control structure, the position controller generates the desired thrust and acceleration, which is tracked by the attitude controller. The force and torque disturbance due to the dynamic coupling of the manipulator is compensated in the position and attitude controller, respectively.
\subsubsection{Dynamics}
In our previous works \cite{zhang_robust_2020}, the whole aerial manipulator is taken as a special aerial platform whose mass distribution is changeable due to the manipulator’s movement. Thus, the disturbance of the  manipulator can be estimated by variable inertia parameters, which are the functions of the manipulator joint state. The dynamics of the quadrotor with manipulator disturbance can be expressed as follows:
\begin{equation}
  \begin{cases}
    \dot{p}_b=v_b                                                                               \\
    \dot{v}_b=-\frac{F}{m_s}R_{IB}  e_3+ge_3+\frac{F_{dis}}{m_s}                                \\
    \dot{R}_{IB}=R_{IB}\cdot skew({}^B\omega _{b})                                            \\
    {}^B\dot{\omega }_b=I_b^{-1}(\tau -{}^{B}\omega _b \times (I_b{}^{B}\omega _b)+\tau _{dis}) \\
  \end{cases}
  \label{eq2}
\end{equation}
where, ${p}_b$ are $v_b$ are the position and velocity of the quadrotor, respectively. $R_{IB}$ is the rotation matrix for frame $\Sigma_B$ to $\Sigma_I$. ${}^B\omega_b$ is the angular velocity of the quadrotor in $\Sigma_B$. $skew(.)$ is skew symmetric matrix function of a vector. $m_s$ is the total mass of the aerial manipulator system. $I_b$ is the inertia matrix of the quadrotor. $F$ and $\tau$ are the thrust and torque generated by the rotors, respectively. $g$ is gravity acceleration. $e_1$, $e_2$ and $e_3$  are unit vectors with 3 dimensions, which means they can form a unit matrix as $I_{3\times3}=\left[ \begin{matrix}  e_1 & e_2 & e_3 \end{matrix} \right]$. $F_{dis}$ and $\tau_{dis}$ are the force and torque disturbance of the manipulator, respectively. The detail expansions of $F_{dis}$ and $\tau_{dis}$ are as follows:
\begin{equation}
  \begin{split}
    F_{dis}= & -m_sR_{IB}({}^B\omega_b \times ({}^B\omega _b \times {}^Br_{com})+{}^B\dot{\omega}_b \times {}^Br_{com} \\
    & +2{}^B{\omega }_b \times {}^B\dot{r}_{com}+{}^B\ddot{r}_{com})
  \end{split}
  \label{eq3}
\end{equation}
\begin{equation}
  \begin{split}
    \tau_{dis}=&-{}^BI_m{}^{B}\dot{\omega}_b-{}^B\omega _b  \times ({}^BI_m{}^B\omega_b)-{}^B I_m{}^B \omega_b \\
    &m_s({}^Br_{com}\times R_{IB}^{-1}(ge_3-\dot{v}_b))-\frac{m_s^{2}}{m_m}{}^Br_{com} \times {}^B\ddot{r}_{com} \\
    &-\frac{m_s^2}{m_m}{}^B\omega_b \times ({}^Br_{com} \times {}^B\dot{r}_{com})
  \end{split}
  \label{eq4}
\end{equation}
where, $m_m$ is the mass of the manipulator. ${}^Br_{com}$ is the position of system CoM (Center of Mass) with respect to $\Sigma_B$. ${}^BI_m{}$ is the inertia matrix the manipulator with respect to $\Sigma_B$. ${}^Br_{com}$ and ${}^BI_m{}$ are the variable inertia parameters, the algorithm to calculate them has been introduced in [14].
\subsubsection{Control law}
First, for the position control, the desired position is denoted by $p_{b,d}$, and the position error is defined as $e_p=p_{b,d}-p_{b}$. The position controller is based on a PID controller with a disturbance compensation term, and its control law is as the following equations: 
\begin{equation}
  \begin{cases}
    a_d=-k_{p,p}e_p-k_{i,p}\int{{e_p}dt}-k_{d,p} \dot{e}_{p}+ge_3+\frac{F_{dis}} {m_s} \\
    F_d=-m_s a_{d}^{T} \cdot R_{IB}e_3                                                \\
  \end{cases}
  \label{eq5}
\end{equation}
where, $k_{p,p}$, $k_{i,p}$and $k_{d,p}$  are the proportional, integral, and derivative gain matrixes of the position PID controller, respectively.  $a_d$ and $F_d$ are the desired acceleration and thrust needed for position control, respectively.

For the attitude control, firstly, the desired rotational matrix should be calculated by the desired acceleration, denoted by $a_d$, and desired yaw angle, denoted by $\psi_d$. 
The desired rotational matrix can be defined as $R_{IB,d}=\left[ \begin{matrix}  x_{B,d} & y_{B,d} & z_{B,d} \end{matrix} \right]^T$, where, $x_{B,d}$, $y_{B,d}$ and $z_{B,d}$ are the desired orientation of frame vectors of $\Sigma_B$ corresponding to $R_{IB,d}$. These frame vectors of $R_{IB,d}$ can be obtained by following equations:
\begin{equation}
  \begin{cases}
    z_{B,d}=\frac{a_d} {\left\| a_d \right\|}                                                      \\
    y_{B,d}=\frac{z_{B,d} \times \tilde{x}_{B,d}} {\left\| z_{B,d}\times \tilde{x}_{B,d} \right\|} \\
    x_{B,d}=y_{B,d} \times z_{B,d}                                                                 \\
  \end{cases}
  \label{eq6}
\end{equation}
where, $\tilde{x}_{B,d}$ is the projection of ${x}_{B,d}$ in the $x\text{-} o \text{-}y$ coordinate plane of $\Sigma_I$, and it has relationship with $\psi_d$ as $\tilde{x}_{B,d}=\left[ \begin{matrix}  \cos\psi_d & \sin\psi_d & 0 \end{matrix} \right]^T$. The attitude error is defined as $e_R=\frac{1}{2}skew^{-1}(R_{IB,d}^{-1}R_{IB}-R_{IB}^{-1}R_{IB,d})$ .The attitude controller is based on a PD controller with a disturbance compensation term, and its control law is as the following equation: 
\begin{equation}
  \tau_d=-k_{p,R}e_R - k_{d,R} \dot{e}_R+\tau_{dis}
  \label{eq7}
\end{equation}
where, $k_{p,R}$ and $k_{d,R}$ are the proportional and derivative gain matrixes of the position PD controller, respectively. $\tau_d$ is the desired torque needed for attitude control

In the strong wind disturbance case, the acceleration feedback based wind disturbance rejection method, which is detailed in our previous work\cite{zhang2019aerial},
can also be applied in this control structure. In this method, acceleration is used to estimate the wind disturbance. Then the estimated disturbance is compensated by the flight controller.
\subsection{Manipulator control}
As shown in Fig. \ref{fig_6}, in the manipulator controller, the endpoint position controller is combined with the inverse velocity kinematics to compensate its position error caused by the float of the quadrotor.
The kinematics of the aerial manipulator endpoint is as follows:
\begin{equation}
  \begin{cases}
    p_{end}=p_b + R_{IB}R_{BM} {}^Mp_{end}\\
    v_{end}=v_b+R_{IB}({}^B\omega _b\times R_{BM}{}^Mp_{end}+R_{BM}{}^Mv_{end}) \\
  \end{cases}
  \label{eq8}
\end{equation}
where, $p_{end}$ and ${}^Mp_{end}$ are the position of manipulator endpoint with respect to $\Sigma_I$ and $\Sigma_M$, respectively. $v_{end}$ and ${}^Mv_{end}$ are the velocity of manipulator endpoint with respect to $\Sigma_I$ and $\Sigma_M$, respectively. $R_{BM}$ is the rotation matrix for frame $\Sigma_M$ to $\Sigma_B$. ${}^Mv_{end}$ can be got through the velocity kinematics of the manipulator. 
The manipulator endpoint position controller is a PD controller which generates the desired velocity of the end, denoted by $v_{end,d}$. With $v_{end,d}$, the inverse kinematics can be used to get desired joint velocity, denoted by $\dot{q}_d$, as follows:
\begin{equation}
  \begin{cases}
    {}^M v_{end,d}=R_{IB}^{-1}(v_{end,d}-v_b)-{}^{B}\omega _b \times {}^Mp_{end} \\
    \dot{q}_d=J(q)^{-1} \cdot {}^Mv_{end,d}                                      \\
  \end{cases}
  \label{eq9}
\end{equation}
where, $q$ is the joint angle of the manipulator. $J(q)$ is the Jacobian matrix of the manipulator. The desired joint velocity $\dot{q}_d$ can be tracked by the actuator's controller.
\subsection{Target pose estimate and hovering set-point}
As introduced in the Sec.II-A, with the special design, the vision system can get images only containing the feature of reflective  markers, as shown in Fig. \ref{fig_3c}. Therefore, it is easy to detect the markers attached in the target robot. With the relative position of the markers and their image coordinates, the target pose can be estimated, which is also known as a PnP (Perspective-n-Point) problem. The method we used to solve the PnP problem is the EPnP algorithm proposed in \cite{penate2013exhaustive}.

With the estimated pose, we can get the hovering set-point that make the target torch at the ideal position in the task workspace. The hovering setpoint and desired yaw angle of the quadrotor, and the desired position of the manipulator endpoint can be obtained as follows:
\begin{equation}
  \begin{cases}
   \psi_d = \psi+^B\tilde{\psi}_t-^B\psi^*_t                          \\ 
    p_{b,d} = p_b + R(\psi,0,0)^B\tilde{p}_t -R(\psi_d,0,0)^Bp^*_t    \\
    p_{end,d} = p_b + R_{IB}\cdot{}^B\tilde{p}_t + \left[0, 0, h_{off}\right]^T 
   \end{cases}
   \label{eq10} 
\end{equation}
where, $^B\tilde{p}_t$ and $^B\tilde{\psi}_t$, are the estimated position and yaw angle of the target torch with respect to $\Sigma_B$, respectively.
$^B\psi^*_t$ and $^Bp^*_t$ are the ideal operating position and yaw angle of the target torch with respect to $\Sigma_B$, respectively. $R(\psi,0,0)$ is the rotation matrix determined by $\psi$, at the condition that the roll and pitch angle are all equal to zero. $p_{end,d}$ is the desired position of the manipulator endpoint with respect to $\Sigma_I$. $h_{off}$ is the height offset of the fire point which is above the end tip of the target torch.

\begin{figure}[H]
  \centering
  \includegraphics[width=1\linewidth]{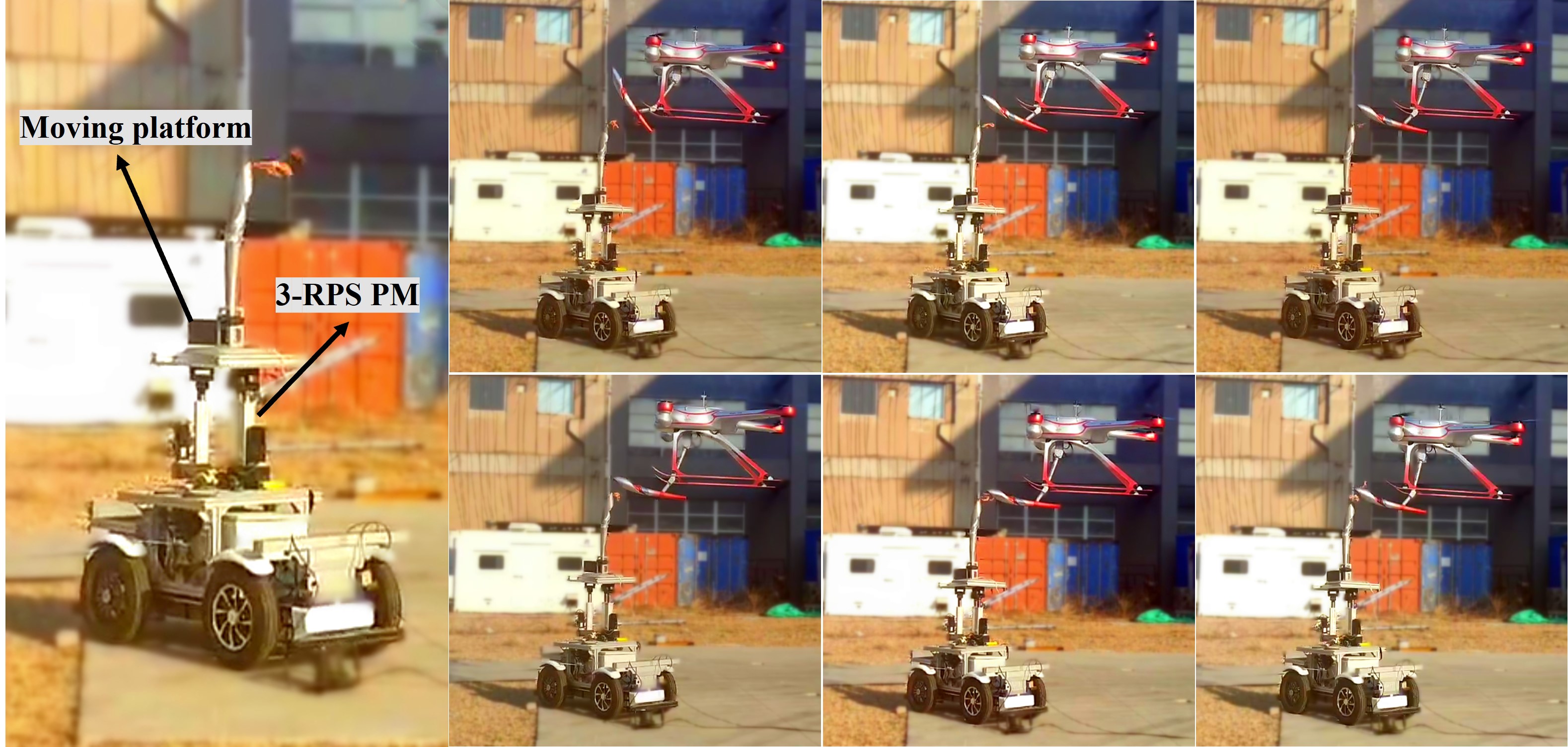}
  \caption{The aerial manipulator performs torch lighting operation with the floating base torch in the ``get fire" case.}
  \label{fig_7}
\end{figure}
\begin{figure*}[t]
  \centering
  \subfloat[]{\includegraphics[width=0.309\linewidth]{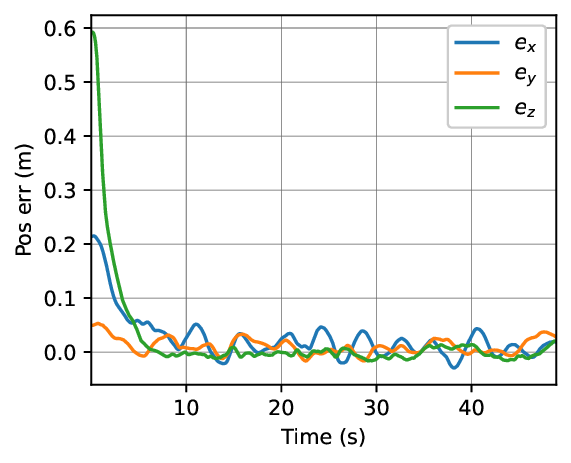}%
    \label{fig10_a}}
  \hfil
  \subfloat[]{\includegraphics[width=0.33\linewidth]{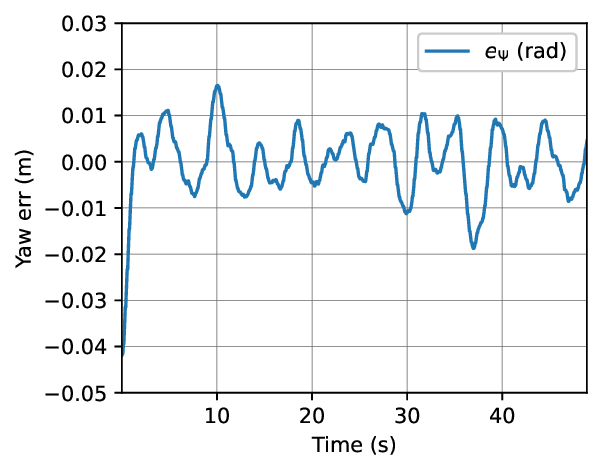}%
    \label{fig10_b}}
  \hfil
  \subfloat[]{\includegraphics[width=0.354\linewidth]{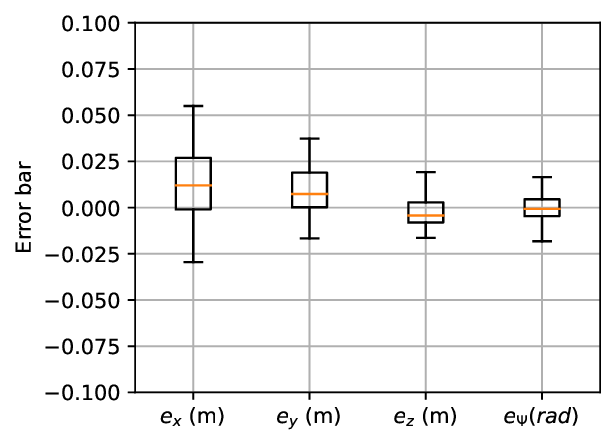}%
  \label{fig10_c}}
  \hfil
  \subfloat[]{\includegraphics[width=0.33\linewidth]{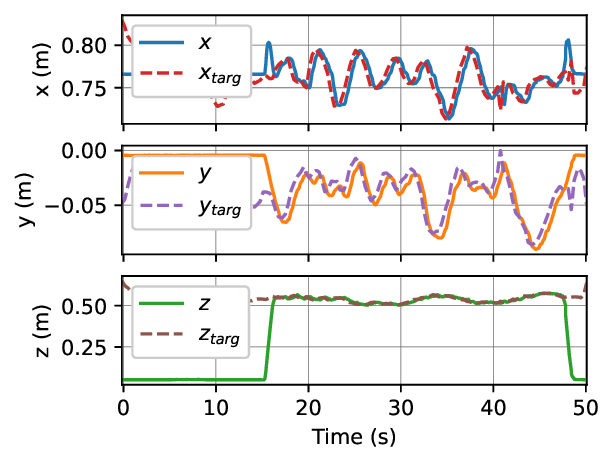}%
  \label{fig10_d}}
  \hfil
  \subfloat[]{\includegraphics[width=0.313\linewidth]{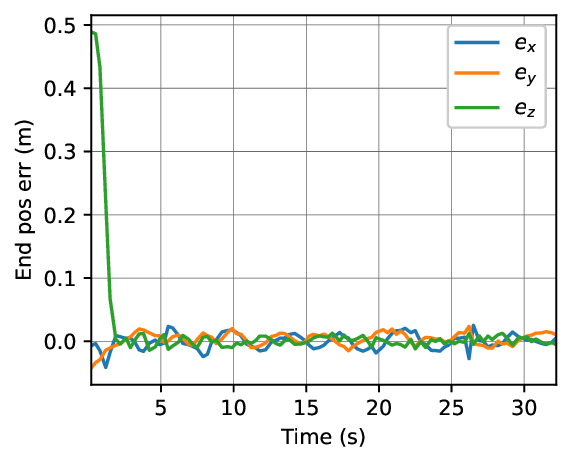}%
  \label{fig10_e}}
  \hfil
  \subfloat[]{\includegraphics[width=0.343\linewidth]{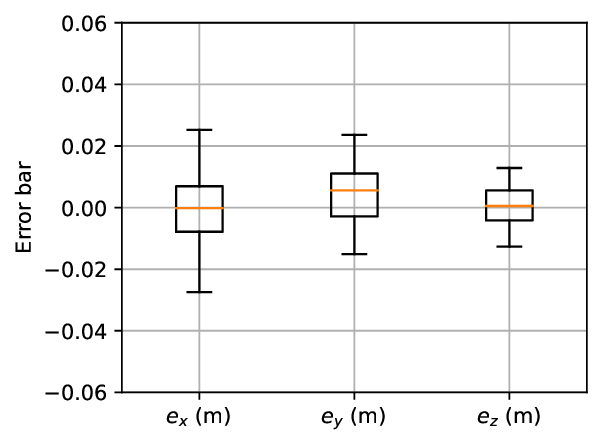}%
  \label{fig10_f}}
  \caption{The experimental results of the aerial manipulator performing the torch lighting operation with a fixed base torch: (a) quadrotor position error; (b) quadrotor yaw error; (c) quadrotor error bar; (d) manipulator endpoint position target tracking; (d) manipulator endpoint position  error; (d) manipulator endpoint position  error bar;}
  \label{fig_10}
\end{figure*}
\begin{figure*}[h]
  \centering
  \subfloat[]{\includegraphics[width=0.325\linewidth]{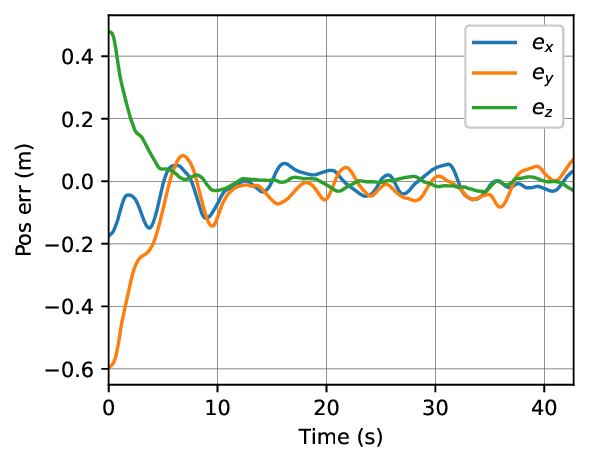}%
    \label{fig9_a}}
  \hfil
  \subfloat[]{\includegraphics[width=0.325\linewidth]{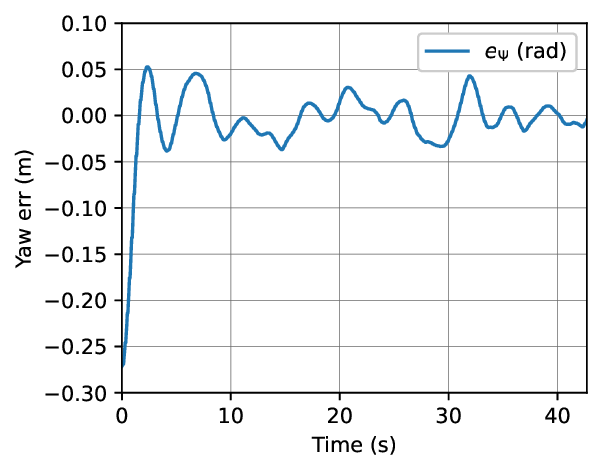}%
    \label{fig9_b}}
  \hfil
  \subfloat[]{\includegraphics[width=0.34\linewidth]{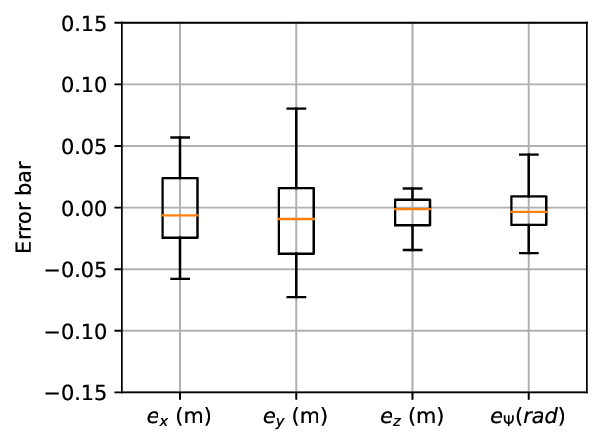}%
  \label{fig9_c}}
  \hfil
  \subfloat[]{\includegraphics[width=0.325\linewidth]{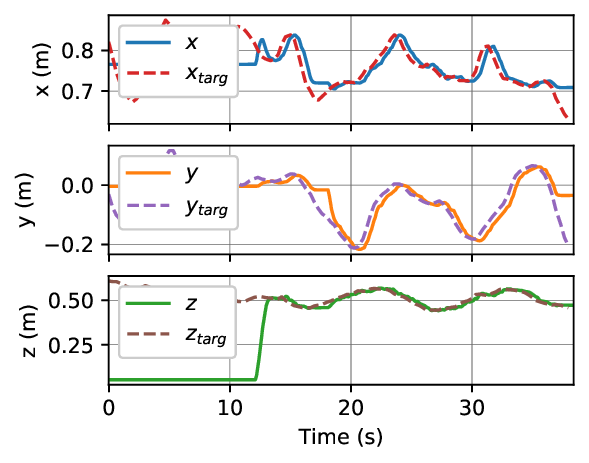}%
  \label{fig9_d}}
  \hfil
  \subfloat[]{\includegraphics[width=0.325\linewidth]{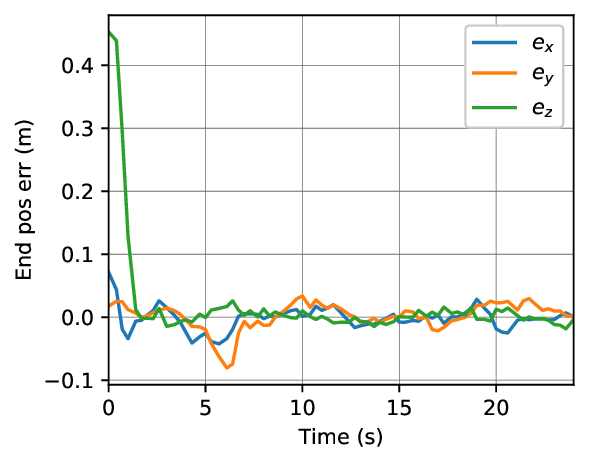}%
  \label{fig9_e}}
  \hfil
  \subfloat[]{\includegraphics[width=0.34\linewidth]{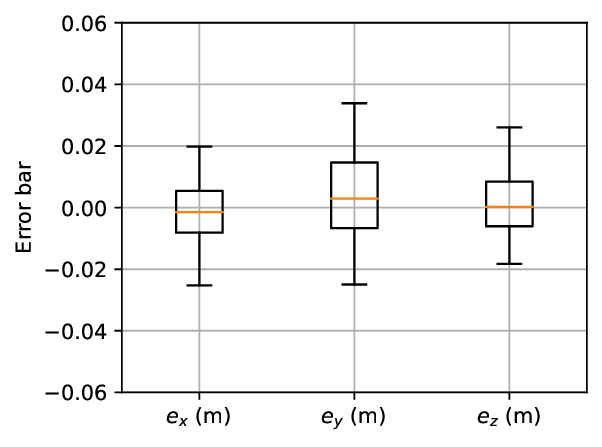}%
  \label{fig9_f}}
  \caption{The experimental results of the aerial manipulator performing the torch lighting operation with a floating base torch :(a) quadrotor position error; (b) quadrotor yaw error; (c) quadrotor error bar; (d) manipulator endpoint position target tracking; (d) manipulator endpoint position error; (d) manipulator endpoint position  error bar;}
  \label{fig_9}
\end{figure*}
\section{Experiments and results}
Several experiments have been conducted to validate that the developed aerial manipulator system has ability to light the torch and can complete the robot-to-robot torch relay task autonomously in the ice and snow field. The experiments and results will be introduced in this section.
\subsection{Experiments of torch lighting}
The aim of the experiments in this subsection is to validate that the aerial manipulator system has ability to light the torch.
\subsubsection{System control accuracy}
Undoubtedly, the control accuracy of the aerial manipulator is a determining factor for the torch lighting task. To validate it, the aerial manipulator would perform the torch lighting operation with a fixed or floating base torch in the experiments. All the experiments were conducted outdoors, and the RTK (Real Time Kinematic) was used to improve the accuracy of GNSS. The torch lighting operation was performed based on visual guidance.
As shown in Fig. \ref{fig_7}, in the floating base torch lighting experiments, the target torch was installed on the moving platform of a 3-RPS (Revolute Prismatic Spherical) parallel manipulator. Because of the parallel mechanical structure, the 3-RPS parallel manipulator can generate a coupled 3D motion of its moving platform. In the experiments, the platform would move up and down, and its attitude would swing along with the movement. For the movement, the trajectory of height is $h_{plat}(t)=0.05\sin(\frac{\pi}{5}t)$. 

The experimental results of the aerial manipulator performing the torch lighting operation with a fixed and floating base torch are shown in Fig. \ref{fig_10} and Fig. \ref{fig_9}, respectively. From the experimental results, we can see that: 1) When the aerial manipulator performs torch lighting operation, it can hover at one point with enough accuracy (within ±10cm) to keep the target torch always in its task workspace, no matter the target torch is floating or not; 2) When the aerial manipulator performs torch lighting operation, the manipulator endpoint can track the target torch with enough accuracy (within ±2cm) to light the torch, even though the target torch is floating with a large range (about 20cm) relative to the aerial manipulator. 
\begin{table}[h]   \small
\caption{The success rate and time average of different cases}
\label{succ_rate_table}
\begin{tabular}{|c|cc|cc|c|}
\hline
\multirow{3}{*}{}               & \multicolumn{2}{c|}{Fixed base}                         & \multicolumn{2}{c|}{Floating base}                       &                       \\ \cline{2-5}
                                & \multicolumn{1}{c|}{success}  & time                    & \multicolumn{1}{c|}{success}   & time                    & $h_{off}$               \\
                                & \multicolumn{1}{c|}{rate}     & agv.                    & \multicolumn{1}{c|}{rate}      & agv.                    &                       \\ \hline
\includegraphics[height=1.5cm]{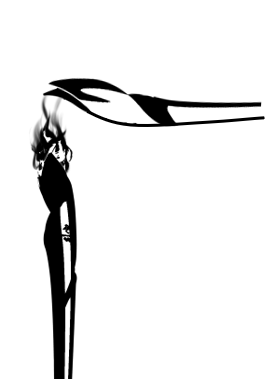}                         & \multicolumn{1}{c|}{5/5}      & \multirow{2}{*}{5.27s}  & \multicolumn{1}{c|}{5/5}       & \multirow{2}{*}{10.23s} & \multirow{2}{*}{10cm} \\
``get fire"                        & \multicolumn{1}{c|}{(100\%)}  &                         & \multicolumn{1}{c|}{(100\%)}   &                         &                       \\ \hline
\includegraphics[height=1.5cm]{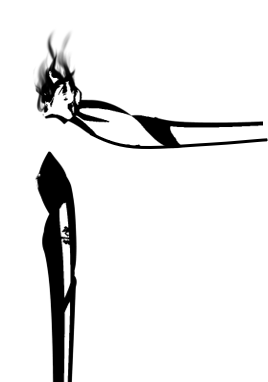}                        & \multicolumn{1}{c|}{8/9}      & \multirow{2}{*}{16.92s} & \multicolumn{1}{c|}{7/11}      & \multirow{2}{*}{24.54s} & \multirow{2}{*}{5cm}  \\
\multicolumn{1}{|l|}{``make fire"} & \multicolumn{1}{l|}{(88.9\%)} &                         & \multicolumn{1}{l|}{(66.64\%)} &                         &                       \\ \hline
\end{tabular}
\end{table}
\subsubsection{Reproducibility}
To evaluate the torch lighting capability in different cases, the torch lighting task is divided into ``get fire"\footnote{To light the aerial manipulator's torch by the target torch} case and ``make fire"\footnote{To light the target torch by the aerial manipulator's torch} case.
Experiments have been conducted several times in the "get fire" and "make fire" case for a fixed and floating base torch lighting task, respectively. The success of the torch lighting task is defined as that the torch is lit within 30 seconds.

The success rate and time average to light the torch of different cases are listed in the \mbox{Table \ref{succ_rate_table}}. From \mbox{Table \ref{succ_rate_table}}, we can see that: 1) In the "make fire" case (with $h_{off}$ equal to 5cm), to light the torch, the manipulator endpoint needs to be closer to the target torch than in the "get fire" case (with $h_{off}$  equal to 10cm); 2) In the "get fire" case, the success rate and efficiency are better. 
The reason is that the gas and flame always float upward, which makes the torch lighting task in the "make fire" case harder than in the "get fire" case. In addition, the level of the remaining gas in the torch and the speed of the wind can also affect the success rate and efficiency of the torch lighting task.
\subsubsection{Results}
Although the aerial manipulator has enough control accuracy to light the torch, other factors should be taken into account to improve the success rate when it is conducting the torch relay task with other robots. If the gas in the torch is full and the Beaufort scale of wind is less than 3, the success rate and efficiency shown in \mbox{Table \ref{succ_rate_table}} can be guaranteed.
\begin{figure}[h]
  \centering
  \subfloat[]{\includegraphics[width=0.35\linewidth]{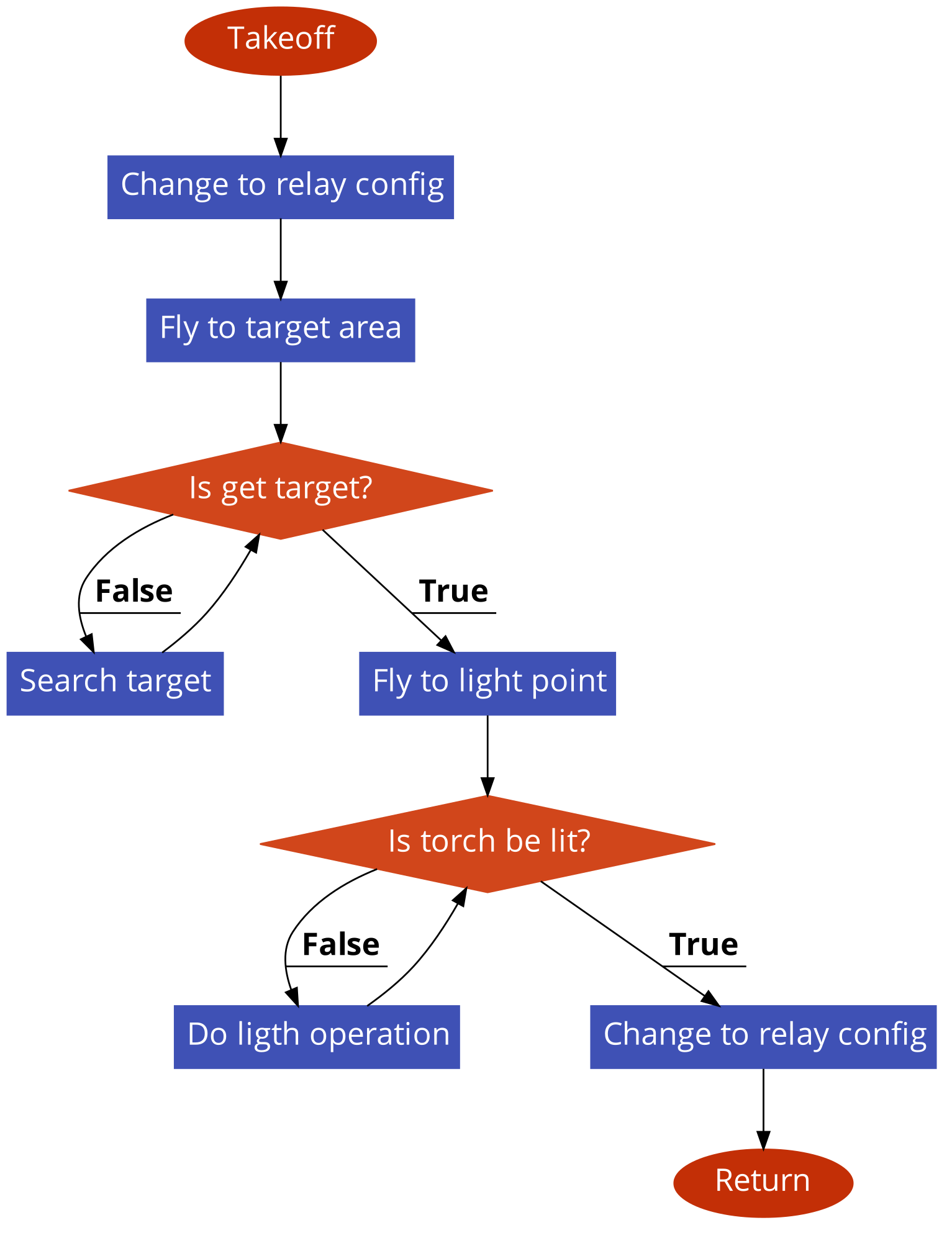} \label{fig8_a}}
  \subfloat[]{\includegraphics[width=0.6\linewidth]{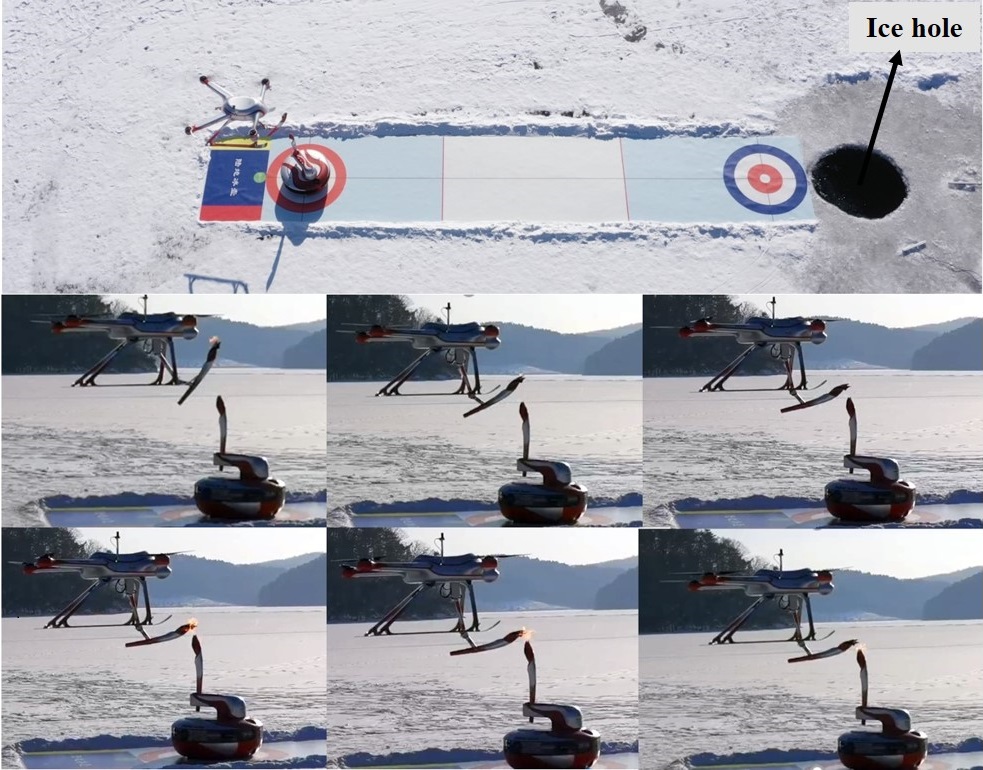} \label{fig8_b}}
  \caption{The aerial manipulator performing the torch relay task with the curling-like amphibious robot: (a) task flow; (b) experimental procedure.}
  \label{fig_8}
\end{figure}
\subsection{Experiments of robot-to-robot torch relay}
In these experiments, the aerial manipulator would perform the torch relay task with the curling-like amphibious robot autonomously in the ice and snow field.
The experiments were conducted in Fushun Dahuofang Reservoir on January 5, 2022. In the torch relay task, the aerial manipulator would light the torch of the curling-like amphibious actively, after which the curling-like amphibious would go into an ice hole to complete the underwater torch relay task.
The task flow of the aerial manipulator is shown in Fig. \ref{fig8_a}.
The experimental procedure is shown in Fig. \ref{fig8_b}.
The experimental videos are available at: \underline{\url{https://youtu.be/ISdDuo3ZBAg}}. 
The experimental results show that the aerial manipulator can complete the torch relay task autonomously during the whole process.

\section{Conclusion}
Towards the robot-to-robot torch relay task of the Beijing 2022 Winter Olympic Games, an aerial manipulator has been  developed. The system design and system control method of the aerial manipulator are detailed in this article. This aerial manipulator system is composed of a quadrotor, a 3 DoF manipulator, and a monocular camera. Based on visual guidance, it can perform the torch lighting operation with enough accuracy. The experimental results demonstrate that it has the ability to complete the torch relay task with other robots autonomously in the ice and snow field.
\section*{Acknowledgments}
This work was supported in part by the National Natural Science Foundation of China Innovation Research Group Project under Grant 61821005 and Shenyang Science and Technology Plan under Grant 21-108-9-18.
\section*{Appendix}
\begin{table}[H]   \small
  \begin{center}
    \caption{Notations}
    \label{tab_notations}
    \begin{tabular}{ l  l}
      \hline
       Symbol          & Description                                \\
      \hline
      $R_{IB}$         & Rotation matrix for frame $\Sigma_B$ to $\Sigma_I$                         \\
      $R_{BM}$         & Rotation matrix for frame $\Sigma_M$ to $\Sigma_B$                         \\
      $p_b$            & Position of the quadrotor\\
      $v_b$            & Velocity of the quadrotor\\
      $^B\omega$       & Angular velocity of the quadrotor in $\Sigma_B$  \\
      $m_s$            & Total mass of the aerial manipulator system.\\
      $m_m$            & Mass of the manipulator. \\
      ${}^Br_{com}$    & Position of the system CoM with respect to $\Sigma_B$. \\
      $^BI_m$          & Inertia matrix of the manipulator with respect to $\Sigma_B$  \\
      $F$              & Total force generated by the rotors\\
      $\tau$           & Total torque generated by the rotors\\
      $F_{dis}$        & Force disturbance of the manipulator\\
      $\tau _{dis}$    & Torque disturbance of the manipulator\\
      $p_{end}$        & Position of the manipulator endpoint with respect \\
                       & to $\Sigma_I$  \\
      $v_{end}$        & Velocity of the manipulator endpoint with respect \\
                       & to $\Sigma_I$  \\
      $^Mp_{end}$      & Position of the manipulator endpoint with respect \\
                       & to $\Sigma_M$  \\
      $^Mv_{end}$      & Velocity of the manipulator endpoint with respect \\
                       & to $\Sigma_M$  \\
      $q$              & Joint angle of the manipulator\\
      $^B\tilde{p}_t$     & Estimated position of the target torch with respect      \\
                          & to $\Sigma_B$  \\
      $^B\tilde{\psi}_t$  & Estimated yaw angle of the target torch with respect    \\
                          & to $\Sigma_B$  \\
      $^B\psi^*_t$        & Ideal operating position of the target torch with    \\
                          & respect to $\Sigma_B$   \\
      $^Bp^*_t$           & Ideal operating yaw angle of the target torch with   \\ 
                          & respect to $\Sigma_B$   \\
      $h_{off}$           & The height offset of the fire point \\
      \hline
    \end{tabular}
  \end{center}
\end{table}


\bibliographystyle{IEEEtran}
\bibliography{IEEEabrv,./bib/BIB_RAM}

\begin{thebibliography}{10}
\providecommand{\url}[1]{#1}
\csname url@samestyle\endcsname
\providecommand{\newblock}{\relax}
\providecommand{\bibinfo}[2]{#2}
\providecommand{\BIBentrySTDinterwordspacing}{\spaceskip=0pt\relax}
\providecommand{\BIBentryALTinterwordstretchfactor}{4}
\providecommand{\BIBentryALTinterwordspacing}{\spaceskip=\fontdimen2\font plus
\BIBentryALTinterwordstretchfactor\fontdimen3\font minus \fontdimen4\font\relax}
\providecommand{\BIBforeignlanguage}[2]{{%
\expandafter\ifx\csname l@#1\endcsname\relax
\typeout{** WARNING: IEEEtran.bst: No hyphenation pattern has been}%
\typeout{** loaded for the language `#1'. Using the pattern for}%
\typeout{** the default language instead.}%
\else
\language=\csname l@#1\endcsname
\fi
#2}}
\providecommand{\BIBdecl}{\relax}
\BIBdecl

\bibitem{gao_robots_2022}
F.~Gao, S.~Li, Y.~Gao, C.~Qi, Q.~Tian, and G.-Z. Yang, ``Robots at the beijing 2022 winter olympics,'' \emph{Sci. Robot.}, vol.~7, no.~65, p. eabq0785, 2022.

\bibitem{ollero_past_2022}
A.~Ollero, M.~Tognon, A.~Suarez, D.~Lee, and A.~Franchi, ``Past, present, and future of aerial robotic manipulators,'' \emph{IEEE Trans. Robot.}, vol.~38, no.~1, pp. 626--645, 2021.

\bibitem{ollero_aeroarms_2018}
A.~Ollero, G.~Heredia, A.~Franchi, G.~Antonelli, K.~Kondak, A.~Sanfeliu, A.~Viguria, J.~R. Martinez-de Dios, F.~Pierri, J.~Cort{\'e}s \emph{et~al.}, ``The aeroarms project: Aerial robots with advanced manipulation capabilities for inspection and maintenance,'' \emph{IEEE Robot. Autom. Mag}, vol.~25, no.~4, pp. 12--23, 2018.

\bibitem{kondak2014aerial}
K.~Kondak, F.~Huber, M.~Schwarzbach, M.~Laiacker, D.~Sommer, M.~Bejar, and A.~Ollero, ``Aerial manipulation robot composed of an autonomous helicopter and a 7 degrees of freedom industrial manipulator,'' in \emph{Proc. 2014 IEEE Int. Conf. Robot. Automat.}\hskip 1em plus 0.5em minus 0.4em\relax IEEE, 2014, pp. 2107--2112.

\bibitem{zhang_grasp_2018}
G.~Zhang, Y.~He, B.~Dai, F.~Gu, L.~Yang, J.~Han, G.~Liu, and J.~Qi, ``Grasp a moving target from the air: System \& control of an aerial manipulator,'' in \emph{Proc. 2018 IEEE Int. Conf. Robot. Automat.}, 2018, pp. 1681--1687.

\bibitem{bartelds_compliant_2016}
T.~Bartelds, A.~Capra, S.~Hamaza, S.~Stramigioli, and M.~Fumagalli, ``Compliant aerial manipulators: Toward a new generation of aerial robotic workers,'' \emph{IEEE Robot. Automat. Lett.}, vol.~1, no.~1, pp. 477--483, 2016.

\bibitem{suarez_lightweight_2018}
A.~Su{\'a}rez, P.~Sanchez-Cuevas, M.~Fernandez, M.~Perez, G.~Heredia, and A.~Ollero, ``Lightweight and compliant long reach aerial manipulator for inspection operations,'' in \emph{Proc. 2018 IEEE/RSJ Int. Conf. Intell. Robots Syst.}, 2018, pp. 6746--6752.

\bibitem{tognon_truly-redundant_2019}
M.~Tognon, H.~A.~T. Ch{\'a}vez, E.~Gasparin, Q.~Sabl{\'e}, D.~Bicego, A.~Mallet, M.~Lany, G.~Santi, B.~Revaz, J.~Cort{\'e}s \emph{et~al.}, ``A truly-redundant aerial manipulator system with application to push-and-slide inspection in industrial plants,'' \emph{IEEE Robot. Automat. Lett.}, vol.~4, no.~2, pp. 1846--1851, 2019.

\bibitem{bodie_dynamic_2021}
K.~Bodie, M.~Tognon, and R.~Siegwart, ``Dynamic end effector tracking with an omnidirectional parallel aerial manipulator,'' \emph{IEEE Robot. Automat. Lett.}, vol.~6, no.~4, pp. 8165--8172, 2021.

\bibitem{chermprayong_integrated_2019}
P.~Chermprayong, K.~Zhang, F.~Xiao, and M.~Kovac, ``An integrated delta manipulator for aerial repair: A new aerial robotic system,'' \emph{IEEE Robot. Autom. Mag}, vol.~26, no.~1, pp. 54--66, 2019.

\bibitem{yang2014dynamics}
H.~Yang and D.~Lee, ``Dynamics and control of quadrotor with robotic manipulator,'' in \emph{Proc. 2014 IEEE Int. Conf. Robot. Automat.}\hskip 1em plus 0.5em minus 0.4em\relax IEEE, 2014, pp. 5544--5549.

\bibitem{Lee2020aerial}
D.~Lee, H.~Seo, D.~Kim, and H.~J. Kim, ``Aerial manipulation using model predictive control for opening a hinged door,'' in \emph{Proc. 20120 IEEE Int. Conf. Robot. Automat.}, 2020, pp. 1237--1242.

\bibitem{chen2022adaptive}
Y.~Chen, J.~Liang, Y.~Wu, Z.~Miao, H.~Zhang, and Y.~Wang, ``Adaptive sliding-mode disturbance observer-based finite-time control for unmanned aerial manipulator with prescribed performance,'' \emph{IEEE Trans. Cybern.}, 2022.

\bibitem{zhang_robust_2020}
G.~Zhang, Y.~He, B.~Dai, F.~Gu, J.~Han, and G.~Liu, ``Robust control of an aerial manipulator based on a variable inertia parameters model,'' \emph{IEEE Trans. Ind. Electron.}, vol.~67, no.~11, pp. 9515--9525, 2019.

\bibitem{tzoumanikas_RSS_2020}
T.~Dimos, G.~Felix, Y.~Qingyue, S.~Dhruv, P.~Marija, and L.~Stefan, ``Aerial manipulation using hybrid force and position nmpc applied to aerial writing,'' in \emph{Proc. of Robot. Sci. Syst.}, 2020.

\bibitem{suarez_physical-virtual_2018}
A.~Suarez, G.~Heredia, and A.~Ollero, ``Physical-virtual impedance control in ultralightweight and compliant dual-arm aerial manipulators,'' \emph{IEEE Robot. Automat. Lett.}, vol.~3, no.~3, pp. 2553--2560, 2018.

\bibitem{lippiello_image-based_2018}
V.~Lippiello, G.~A. Fontanelli, and F.~Ruggiero, ``Image-based visual-impedance control of a dual-arm aerial manipulator,'' \emph{IEEE Robot. Automat. Lett.}, vol.~3, no.~3, pp. 1856--1863, 2018.

\bibitem{hu_vision-based_2022}
A.~Hu, M.~Xu, H.~Wang, and H.~Casta{\~n}eda, ``Vision-based impedance control of an aerial manipulator using a nonlinear observer,'' \emph{IEEE Trans. Autom. Sci. Eng.}, 2022.

\bibitem{zhang2019aerial}
G.~Zhang, Y.~He, B.~Dai, F.~Gu, L.~Yang, J.~Han, and G.~Liu, ``Aerial grasping of an object in the strong wind: Robust control of an aerial manipulator,'' \emph{Appl. Sci.}, vol.~9, no.~11, p. 2230, 2019.

\bibitem{penate2013exhaustive}
A.~Penate-Sanchez, J.~Andrade-Cetto, and F.~Moreno-Noguer, ``Exhaustive linearization for robust camera pose and focal length estimation,'' \emph{IEEE Trans. Pattern Anal. Mach. Intell.}, vol.~35, no.~10, pp. 2387--2400, 2013.

\end{thebibliography}


\end{document}